\documentclass{article}
\usepackage{spconf,amsmath,graphicx}
\usepackage{multirow}
\usepackage{xcolor}
\usepackage{amssymb}
\usepackage{marvosym}

\title{I-Tuning: Tuning Frozen Language Models with Image for\\Lightweight Image Captioning}
\name{Ziyang Luo$^{\star}$ \qquad Zhipeng Hu$^{\dagger\ddagger}$ \qquad Yadong Xi$^{\ddagger}$ \qquad Rongsheng Zhang$^{\ddagger}$ \qquad Jing Ma$^{\star}\textsuperscript{\Letter}$\thanks{\Letter: Corresponding author (majing@hkbu.edu.hk)}}
\address{$^{\star}$ Department of Computer Science, Hong Kong Baptist University, Hong Kong SAR, China \\
$^{\dagger}$ College of Computer Science and Technology, Zhejiang University, Hangzhou, China \\
$^{\ddagger}$ Fuxi AI Lab, NetEase Inc., Hangzhou, China}

\begin{document}
%\ninept
%
\maketitle
\begin{abstract}

Image Captioning is a traditional vision-and-language task that aims to generate the language description of an image. Recent studies focus on scaling up the model size and the number of training data, which significantly increase the cost of model training. Different to these heavy-cost models, we introduce a lightweight image captioning framework (\textbf{I-Tuning}), which contains a small number of trainable parameters. We design a novel I-Tuning cross-attention module to connect the non-trainable pre-trained language decoder GPT2 and vision encoder CLIP-ViT. Since most parameters are not required to be updated during training, our framework is lightweight and fast. Experimental results conducted on three image captioning benchmarks reveal that our framework achieves comparable or better performance than the large-scale baseline systems. But our models contain up to \textbf{10 times fewer} trainable parameters and require much fewer data for training compared with state-of-the-art baselines.

\end{abstract}
\section{Introduction}

Image Captioning is a critical task in the field of cross-modal, which focus on natural language generation to depict an image. Recent years have witnessed the success of applying large-scale pre-trained models on the task of image captioning, which generally scale up the number of trainable parameters and training data to achieve state-of-the-art performances~\cite{li2020oscar,zhou2019unified,xia2020xgpt,yang2021crossing}. For example, a recent proposed OSCAR model~\cite{li2020oscar} contains more than 135M trainable parameters and requires around 4M images during pre-training. Therefore, in spite of the performances, the heavy demands for extra computational resources and massive data for model training have become an urgent issue. %is a non-negligible issue. 
 
\begin{figure}
    \centering
    \includegraphics[height=3.7cm]{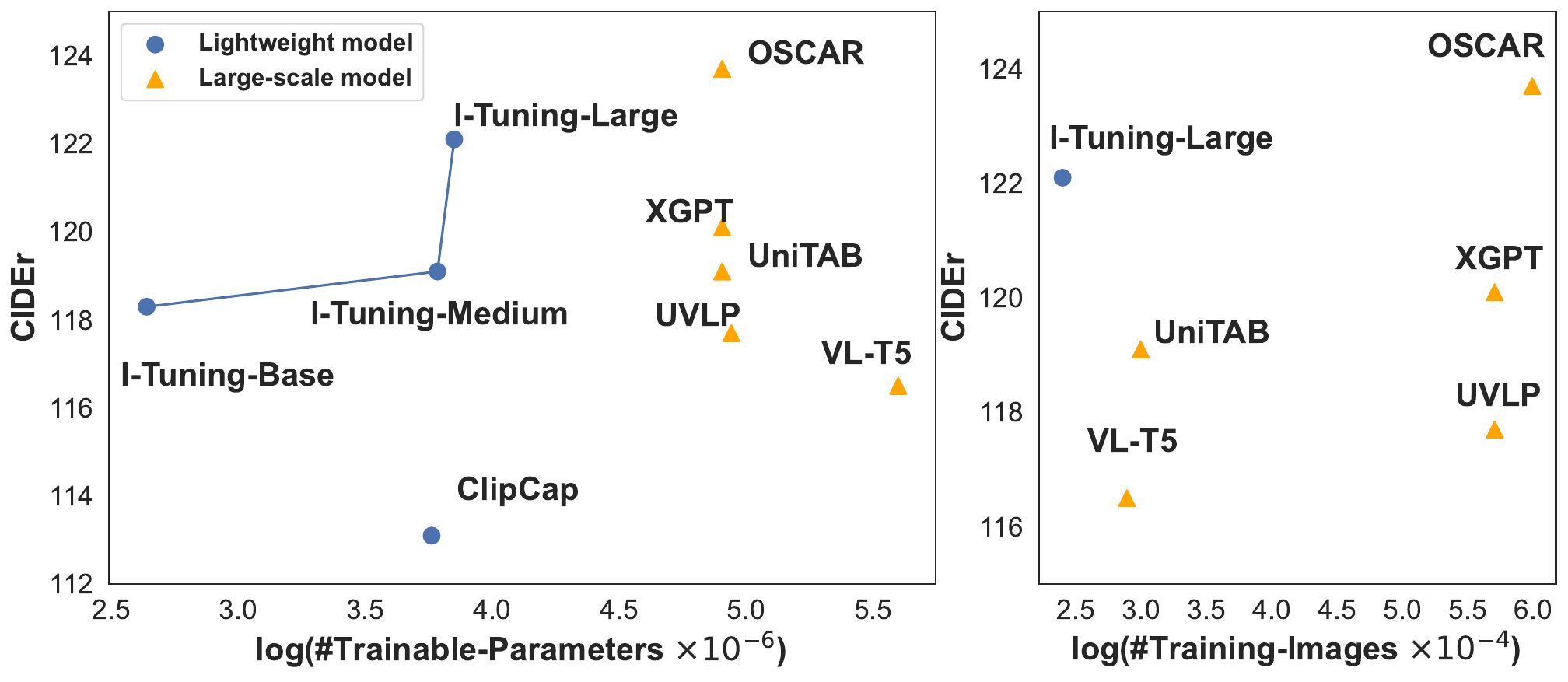}
    \caption{Image captioning performances on MSCOCO dataset based on the lightweight and large-scale models, where I-Tuning-Base/-Medium/-Large respectively denote our model with base/medium/large number of trainable parameters.} %  with different number of trainable parameters.}
    \label{fig:compare}
\end{figure}

Recent studies showed that parameter-efficient pre-trained language models (PLMs) tuning~\cite{pmlr-v97-houlsby19a} can effectively reduce the cost during training, where most parameters are frozen (i.e., not updated during training) and the rest small set are trainable. \cite{mokady2021clipcap} recently introduced a ClipCap model that transforms images into fixed-length vectors and prompts a frozen GPT2 for image captioning. However, the learned vectors cannot capture accurate visual information to enhance the caption generation.
%
%As a result, less trainable parameters lead to less backward computations and faster training speed. Following this paradigm, 
To overcome these shortcomings, we propose a novel lightweight image captioning framework (\textbf{I-Tuning}) to alleviate the cost in terms of computational resource and training data. %introduces \textbf{I-Tuning, our lightweight image captioning framework} as an alternative to the heavy-cost image captioning models. 
We design an I-Tuning module to connect the pre-trained vision encoder (i.e., CLIP-ViT~\cite{radford2021learning}) and the language decoder (i.e., GPT2~\cite{radford2019language}). To align between the language and vision modals, it serves as a cross-modal filter that automatically picks the visual information from the output of the vision encoder and adjusts the output hidden states of the language decoder. During training, we only update the newly introduced parameters in the I-Tuning module, and the parameters of the two pre-trained models are frozen. 

Figure~\ref{fig:compare} examplifies the CIDEr scores of our lightweight models and large-scale baselines. In terms of model training, our basic model I-Tuning-Base %With this design, our model is lightweight and the smallest one that 
only contains around 14M trainable parameters, namely \textbf{10 times fewer} than the other large-scale models such as OSCAR. In terms of data, even our I-Tuning-Large model can achieve comparable performances with relatively less training data. %Moreover, the off-the-shelf pre-trained models can naturally alleviate the dependent on large-scale data for training visual recognition and language generation model. %Therefore, our framework is lightweight for computational cost and training data requirements.
We evaluate our proposed framework on 3 image captioning benchmarks (i.e., MSCOCO~\cite{lin2015microsoft}, Flickr30k~\cite{plummer2016flickr30k} and NoCaps~\cite{nocaps_paper}). The results show that our \textbf{I-Tuning} framework achieves comparable or even better performances than large-scale baselines with up to \textbf{10 times fewer} trainable parameters and much fewer cross-modal training data. Moreover, our I-Tuning model is agnostic to the pre-trained language models, suggesting a broadly applicable framework. % In addition, we show that \textbf{I-Tuning} are agnostic to the pre-trained language models, which makes our method broadly applicable.

\section{Related Work}

\textbf{CLIP-ViT and GPT2.} CLIP-ViT~\cite{radford2021learning} is the state-of-the-art vision encoder. It is pre-trained with contrastive loss~\cite{He2020MomentumCF} to supervise the vision encoder with language description. GPT2~\cite{radford2019language} is the state-of-the-art language decoder, which is pre-trained with large-scale text data. In this work, we propose a lightweight image captioning framework \textbf{I-Tuning} to leverage these two off-the-shelf pre-trained models.

\textbf{Image Captioning.} Generating the language descriptions from images is an important task to examine the vision-and-language representation ability of a cross-modal model. The recent works choose to increase the model size and the number of training data to further boost the performance~\cite{li2020oscar,zhou2019unified,xia2020xgpt,yang2021crossing,wang2022simvlm}. The training process of these models is heavy. As an alternative, the ClipCap model~\cite{mokady2021clipcap} proposes a lightweight captioning model by connecting the off-the-shelf CLIP-ViT and GPT2. However, their method cannot filter the relevant visual information to adjust the output hidden states of GPT2, leading to poor image captioning performance. %We propose the \textbf{I-Tuning} module to fill this gap.

\textbf{Parameter-efficient PLMs Tuning.} Recently, the model size of a pre-trained model becomes larger and larger, which makes us hard to fully fine-tune such models. To make use of them without updating all parameters, researchers propose several great ideas, such as Prefix tuning~\cite{li-liang-2021-prefix}, Adapter tuning~\cite{pmlr-v97-houlsby19a} and Prompt tuning~\cite{liu2021gpt}. However, most of them only focus on the NLP area. Our \textbf{I-Tuning} extends the parameter-efficient PLMs tuning idea to the cross-modal setting.

\section{The Proposed I-Tuning Framework}

\textbf{Overview.} Our framework contains three components, the non-trainable vision encoder (CLIP-ViT), the non-trainable language decoder (GPT2), and the trainable I-Tuning Module. During training, our framework is trained with the parallel image-caption data and only updated the parameters of the lightweight I-Tuning Module.

During inference, a frozen visual encoder first generates the visual embeddings $V$ of a given image. Then the \textbf{I-Tuning} module serves as a lightweight filter to pick the relevant visual information to tune the output hidden states of the frozen language model. As a result, the language generation is conditioned with the given image. 

\begin{figure}
    \centering
    \includegraphics[height=5cm]{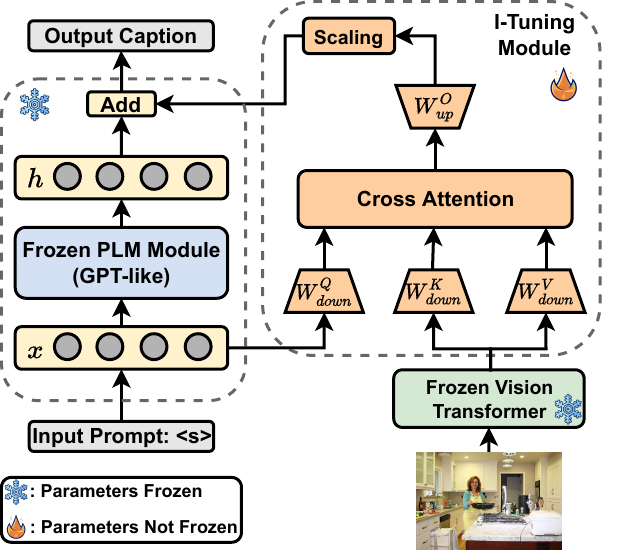}
    \caption{\textbf{An overview of our I-Tuning framework for Lightweight Image Captioning.}}
    \label{fig:overview}
\end{figure}

\textbf{Visual Encoder and Language Decoder.} In our framework, we adopt the state-of-the-art vision pre-trained transformer, CLIP-ViT~\cite{radford2021learning} to generate an image's visual embeddings $V$. Such model takes a sequence of image patches as input and visual representations for each patch as output. For the Language Decoder, we leverage the state-of-the-art auto-regressive pre-trained language model (PLM), GPT2~\cite{radford2019language}, which is a multi-layer Transformer Decoder model~\cite{NIPS2017_3f5ee243} with remarkable language generation ability.

\begin{table*}[h!]
    \footnotesize
    \centering
    \begin{tabular}{l|c|c|cccc|cccc}
        \hline
        \multirow{2}{*}{\textbf{Model}} & \multirow{2}{*}{\textbf{\#Images}} & \multirow{2}{*}{\textbf{\#Params}} & \multicolumn{4}{c|}{\textbf{MSCOCO (test)}} & \multicolumn{4}{c}{\textbf{Flickr (test)}}\\
        & & & \textbf{CIDEr} & \textbf{BLUE@4} & \textbf{METER} & \textbf{SPICE} & \textbf{CIDEr} & \textbf{BLUE@4} & \textbf{METER} & \textbf{SPICE}\\
        \hline
        \hline
        \multicolumn{9}{l}{\textit{Large-scale Cross-Modal Pre-trained Image Captioning Model}}\\
        \hline
        \textbf{OSCAR$_{base}$}(no tags)~\cite{li2020oscar} & 4M & 135M & 115.6 & 34.5 & 29.1 & 21.9 & - & - & - & -\\
        \textbf{OSCAR$_{base}$}$\spadesuit$~\cite{li2020oscar} & 4M & 135M & 123.7 & 36.5 & 30.3 & 23.1 & - & - & - & -\\
        \textbf{Unified VLP}~\cite{zhou2019unified} & 3M & 135M & 117.7 & 36.5 & 28.4 & 21.3 & 67.4 & 30.1 & 23.0 & 17.0\\
        \textbf{XGPT}~\cite{xia2020xgpt} & 3M & 135M & 120.1 & 37.2 & 28.6 & 21.8 & 70.9 & 31.8 & 23.6 & 17.6\\
        \textbf{UniTAB}~\cite{yang2021crossing} & 200k & 135M & 119.1 & 35.8 & 28.4 & 21.5 & 70.1 & 30.7 & 23.7 & 17.4\\
        \textbf{VL-T5}~\cite{pmlr-v139-cho21a} & 180k & 270M & 116.5 & 34.5 & 28.7 & 21.9 & - & - & - & -\\
        \hline
        \hline
        \multicolumn{9}{l}{\textit{Lightweight Image Captioning Model}}\\
        \hline
        \textbf{ClipCap}(GPT2-Large)~\cite{mokady2021clipcap} & 0 & 43M & 113.1 & 33.5 & 27.5 & 21.1 & - & - & - & -\\
        \hline
        \multicolumn{9}{l}{\textit{Our Lightweight Models w/o VLP}}\\
        \hline
        \textbf{I-Tuning}(GPT2-Base) & 0 & 14M & 116.7 & 34.8 & 28.3 & 21.8 & 61.5 & 25.2 & 22.8 & 16.9\\
        \textbf{I-Tuning}(GPT2-Medium) & 0 & 44M & 120.0 & 35.5 & 28.8 & 22.0 & 72.3 & 28.8 & 24.6 & 19.0\\
        \textbf{I-Tuning}(GPT2-Large) & 0 & 95M & 119.4 & 34.8 & 29.3 & 22.4 & 75.4 & 29.8 & 25.1 & 19.2\\
        \hline
        \multicolumn{9}{l}{\textit{Our Lightweight Models w/ VLP}}\\
        \hline
        \textbf{I-Tuning}(GPT2-Base) & 110k & 14M & 118.3 & 35.2 & 28.5 & 22.0 & 68.4 & 27.5 & 24.0 & 18.4\\
        \textbf{I-Tuning}(GPT2-Medium) & 110k & 44M & 119.1 & 34.8 & 29.2 & 22.2 & 73.2 & 29.1 & 25.2 & 19.9\\
        \textbf{I-Tuning}(GPT2-Large) & 110k & 95M & \textbf{122.2} & 35.9 & \textbf{29.5} & \textbf{22.6} & 77.2 & 30.0 & \textbf{25.5} & \textbf{20.2}\\
        \hline
        \multicolumn{9}{l}{\textit{Our Lightweight Models w/ I-Tuning Dropping}}\\
        \hline
        \textbf{I-Tuning}(GPT2-Large) & 110k & 47M & 122.1 & \textbf{36.1} & 29.4 & \textbf{22.6} & \textbf{79.2} & \textbf{31.1} & 25.3 & 19.9\\
        \hline
    \end{tabular}
    \caption{Evaluations on MSCOCO and Flickr Image Captioning. ``-'' represents that the model does not report such result in its original paper. \textbf{Bold} indicates the best scores of our models. \underline{\#Images} represents the number of distinct images during VLP.  \underline{\#Params} represents the number of trainable parameters. $\spadesuit$: Extra training data are needed to generate the object tags.)}
    \label{tab:previous_work_comparsion}
\end{table*}

\textbf{I-Tuning Module.} In our framework, the \textbf{I-Tuning} module is the key component to extract the relevant visual information from the visual embeddings, which is parallel to a specific PLM module (feedforward) in each Transformer layer. Such module is a bottleneck neural network, sharing a similar structure as the Adapter module~\cite{pmlr-v97-houlsby19a}, but the non-linear activation function is replaced by a cross-attention network (see Figure~\ref{fig:overview}) to filter the visual information from images. The calculation process is as follows:
\begin{align}
    Q_{L} &= W_{down}^Q(X) + b^Q,\\
    K_{V} &= W_{down}^K(V) + b^K,\\
    V_{V} &= W_{down}^V(V) + b^V,
\end{align}
where $X$ is the input hidden states of a specific PLM module. Then we can get the attention scores across the visual embeddings:
\begin{equation}
    S = softmax(Q_LK^T_V).
\end{equation}
Based on the scores, we can get the final \textbf{I-Tuning} output to adjust the output hidden states of the PLM module:
\begin{equation}
    \Delta h = \lambda W_{up}^O\left(\sum_{i}s_iV_{Vi}\right)+b^O,
\end{equation}
where $\lambda\geq 1$ is a scaling hyper-parameter.

Since the lower layers of PLMs have weaker representation ability, we also propose \textbf{I-Tuning Dropping} to remove the \textbf{I-Tuning} modules in the first-few layers. As a result, backpropagating through fewer layers can further improve the training efficiency of our models.

\textbf{Training Objective.} The objective is the auto-regressive language modeling conditioned on the visual information: $\mathcal{L}=-\Sigma_{t=1}^T\log P(x_t|x_{<t}, V)$,
where $V$ represents the visual embeddings encoded by the frozen visual encoder, T denotes the length of a sequence and $x_{<t} = (x_0,...,x_{t-1})$.

\section{Experiment}

\subsection{Dataset and Setup}

We adopts CLIP-ViT B/16 as our visual encoder and GPT2 as language decoder. All of them are frozen during training. We include 3 different GPT2 model sizes, including Base, Medium and Large. For \textbf{I-Tuning} modules, the parameters are randomly initialized and updated during training. For VLP, we adopt the cross-modal dataset, Visual Genome~\cite{Krishna2016VisualGC}, which contains 110k distinct images. To evaluate our methods, we use three datasets, namely \textbf{MSCOCO}~\cite{lin2015microsoft}, \textbf{Flickr30k}~\cite{plummer2016flickr30k} and \textbf{NoCaps}~\cite{nocaps_paper}. For the first two datasets, we follow the Karpathy’s split~\cite{karpathy2015deep} to split 113.2k/5k/5k and 29.8k/1k/1k images for train/val/test, respectively. We adopt  CIDEr~\cite{vedantam2015cider}, BLEU@4~\cite{papineni-etal-2002-bleu}, METEOR~\cite{banerjee-lavie-2005-meteor} and SPICE~\cite{anderson2016spice} as metrics to evaluate the generated captions. We train our models with the AdamW~\cite{loshchilov2019decoupled} and 4k batch size. For VLP, our models are pre-trained with 10 epochs. For training on downstream tasks, our models are trained with 30 epochs. For inference, we use the beam search (beam size = 5) to generate captions.

\begin{table*}[h!]
    \footnotesize
    \centering
    \begin{tabular}{l|c|cc|cc|cc|cc}
        \hline
        \multicolumn{1}{l|}{\multirow{2}{*}{\textbf{Model}}} & \multirow{2}{*}{\textbf{\#Params}} & \multicolumn{2}{c|}{\textbf{in-domain}} & \multicolumn{2}{c|}{\textbf{near-domain}} & \multicolumn{2}{c|}{\textbf{out-of-domain}} & \multicolumn{2}{c}{\textbf{Overall}}     \\
        \multicolumn{1}{l|}{} & \multicolumn{1}{l|}{} & \textbf{CIDEr} & \multicolumn{1}{c|}{\textbf{SPICE}} & \textbf{CIDEr} & \multicolumn{1}{c|}{\textbf{SPICE}} & \textbf{CIDEr} & \multicolumn{1}{c|}{\textbf{SPICE}} & \textbf{CIDEr} & \multicolumn{1}{c}{\textbf{SPICE}}\\
        \hline
        \hline
        \textbf{OSCAR}$_{base}$~\cite{li2020oscar} & 135M & 79.6 & 12.3 & 66.1 & 11.5 & 45.3 & 9.7 & 63.8 & 11.2\\
        \textbf{ClipCap}(GPT2-Large)~\cite{mokady2021clipcap} & 43M & 84.9 & 12.1 & 66.8 & 10.9 & 49.1 & 9.6 & 65.8 & 10.9\\
        \hline
        \hline
        \multicolumn{9}{l}{\textit{Our Models}}\\
        \hline
        \textbf{I-Tuning}(GPT2-Base) & 14M & 83.9 & 12.4 & 70.3 & 11.7 & 48.1 & 9.5 & 67.8 & 11.4\\
        \textbf{I-Tuning}(GPT2-Medium) & 44M & \textbf{89.6} & 12.9 & 77.4 & 12.2 & 58.8 & 10.5 & 75.4 & 12.0\\
        \textbf{I-Tuning}(GPT2-Large) & 95M & \textbf{89.6} & \textbf{13.3} & 80.4 & \textbf{12.6} & 64.8 & \textbf{11.0} & 78.5 & \textbf{12.4}\\
        \hline
        \multicolumn{9}{l}{\textit{Our Models w/ Dropping}}\\
        \hline
        \textbf{I-Tuning}(GPT2-Large) & 47M & 88.3 & 12.7 & \textbf{80.8} & \textbf{12.6} & \textbf{66.1} & 10.8 & \textbf{78.9} & 12.3\\
        \hline
    \end{tabular}
    \caption{Evaluations on NoCaps image captioning. Models are only trained with MSCOCO training set without VLP. \underline{\#Params} represents the number of trainable parameters. \textbf{Bold} indicates the best scores.}
    \label{tab:nocaps}
\end{table*}

\subsection{Result Analysis}

Table~\ref{tab:previous_work_comparsion}--\ref{tab:nocaps} reveal that our lightweight image captioning framework achieves comparable or better performance than all the large-scale baselines, but contains up to 10 times fewer trainable parameters and/or consume much fewer VLP data. 

\textbf{I-Tuning without VLP.} As shown in Table~\ref{tab:nocaps}, our method outperform the large-scale baselines even without VLP. Especially, the overall CIDEr score of the OSCAR model on the NoCaps even lags behind the frozen GPT2-base with our \textbf{I-Tuning} modules by around 2 points, while our model contains around 120M fewer trainable parameters. With the larger GPT2, the performance gap becomes larger.
Moreover, our method is also sample efficient. Without any cross-modal pre-training, our \textbf{I-Tuning} (GPT2-Medium) already outperforms some baselines with VLP. For example, VL-T5 is pre-trained with 180k distinct cross-modal images, but the CIDEr scores are around 3.4 lower than ours on MSCOCO.

\begin{table*}[h!]
    \scriptsize
    \centering
    \begin{tabular}{l|llll}
        \hline
        \textbf{Image} &
         \begin{minipage}[b]{0.4\columnwidth}
		    \centering
		    \raisebox{-.5\height}{\includegraphics[width=0.6\linewidth]{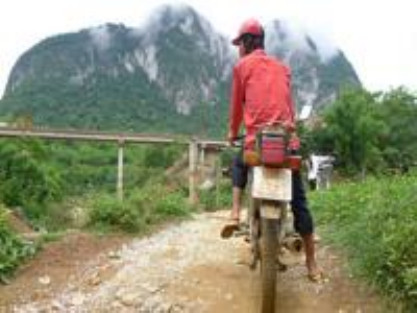}}
	    \end{minipage}
	    &
	    \begin{minipage}[b]{0.4\columnwidth}
		    \centering
		    \raisebox{-.5\height}{\includegraphics[width=0.6\linewidth]{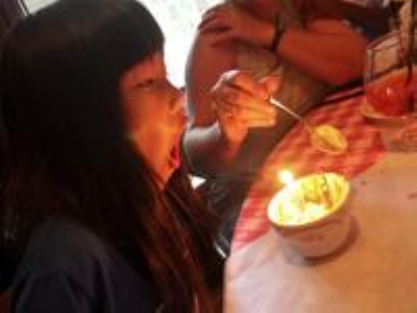}}
	    \end{minipage}
	    &
	    \begin{minipage}[b]{0.4\columnwidth}
		    \centering
		    \raisebox{-.5\height}{\includegraphics[width=0.6\linewidth]{coco3.pdf}}
	    \end{minipage}
	    &
	    \begin{minipage}[b]{0.4\columnwidth}
		    \centering
		    \raisebox{-.5\height}{\includegraphics[width=0.6\linewidth]{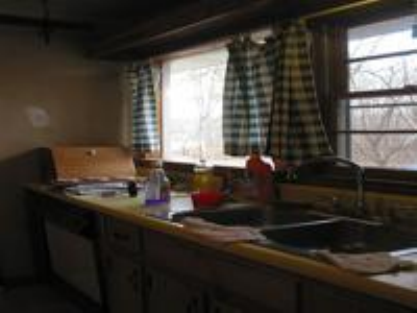}}
	    \end{minipage}
         \\
         \hline
         \hline
        \begin{tabular}[c]{@{}l@{}}
            \textbf{Golden}\\
            \textbf{Captions}\\
        \end{tabular}
        & 
        \begin{tabular}[c]{@{}l@{}}
            (1) A man with a red helmet on a\\
            small moped on a dirt road.\\
            (2) Man riding a motor bike on a\\
            dirt road on the countryside.\\
        \end{tabular}
        &
        \begin{tabular}[c]{@{}l@{}}
            (1) A young girl inhales with the\\
            intent of blowing out a candle.\\
            (2) A young girl is preparing to\\
            blow out her candle.\\
        \end{tabular}
        &
         \begin{tabular}[c]{@{}l@{}}
            (1) A man on a bicycle riding next\\
            to a train.\\
            (2) A person is riding a bicycle but\\
            there is a train in the background.\\
        \end{tabular}
        &
        \begin{tabular}[c]{@{}l@{}}
            (1) A kitchen is shown with a variety\\
            of items on the counters.\\
            (2) A kitchen has the windows open\\
            and plaid curtains.\\
        \end{tabular}
        \\
        \hline
        \hline
        \textbf{Model} &\multicolumn{4}{c}{\textbf{Generated Caption}}\\
        \hline
        \begin{tabular}[c]{@{}l@{}}
            \textbf{ClipCap}\\
            (GPT2-Large)\\
        \end{tabular}
        &
        \begin{tabular}[c]{@{}l@{}}
            a man is riding a motorbike on\\
            a dirt road.\\
        \end{tabular}
        &
        \begin{tabular}[c]{@{}l@{}}
            a young girl {\color{red}sitting} at a table\\
            with a cup of cake.\\
        \end{tabular}
        &
        \begin{tabular}[c]{@{}l@{}}
            a man is {\color{red}standing} next to a train.
        \end{tabular}
        &
        \begin{tabular}[c]{@{}l@{}}
            a kitchen with {\color{red}a} sink and {\color{red}a} window\\
        \end{tabular}
        \\
        \hline
        \textbf{OSCAR}$_{base}$
        &
        \begin{tabular}[c]{@{}l@{}}
            a man riding a motorcycle down\\
            a dirt road.\\
        \end{tabular}
        &
        \begin{tabular}[c]{@{}l@{}}
            {\color{red} a woman sitting at a table with}\\
            {\color{red} a plate of food}.
        \end{tabular}
        &
        \begin{tabular}[c]{@{}l@{}}
            a {\color{red} woman} riding a bike down a\\
            street next to a train.\\
        \end{tabular}
        &
        \begin{tabular}[c]{@{}l@{}}
            a kitchen with {\color{red}a} sink, dishwasher\\
            and {\color{red}a} window\\
        \end{tabular}
        \\
        \hline
        \begin{tabular}[c]{@{}l@{}}
            \textbf{I-Tuning}\\
            (GPT2-Large)\\
        \end{tabular}
        &
        \begin{tabular}[c]{@{}l@{}}
            A man riding a motorcycle on\\
            a dirt road.\\
        \end{tabular}
        &
        \begin{tabular}[c]{@{}l@{}}
            A little girl blowing out a\\
            candle on a birthday cake.\\
        \end{tabular}
        &
        \begin{tabular}[c]{@{}l@{}}
            A man {\color{red}standing} next to a train\\
            on a train track.\\
        \end{tabular}
        &
        \begin{tabular}[c]{@{}l@{}}
            A kitchen sink sitting under a window\\
            next to a window.\\
        \end{tabular}
        \\
        \hline
    \end{tabular}
    \caption{Examples of our \textbf{I-Tuning}, OSCAR$_{base}$ and ClipCap for the first 4 images in the MSCOCO test set. ({\color{red} Red} = inaccurate)}
    \label{tab:generated_example}
\end{table*}

\textbf{I-Tuning with VLP.} Table~\ref{tab:previous_work_comparsion} reveals that after cross-modal pre-training, our \textbf{I-Tuning} method achieves better overall performance than all the baseline systems (except OSCAR w/ object tags). Especially, our \textbf{I-Tuning} can achieve a CIDEr score of 122.1 on MSCOCO test set, surpassing the XGPT model by 2.0 points, while our method requires less trainable parameters and training data. For the OSCAR model, it requires object tags during pre-training and fine-tuning. Additional supervision is needed to generate these tags. One can find that our lightweight \textbf{I-Tuning} framework still reaches comparable performance with only 1.5 CIDEr score lower, while our model requires around 90M less trainable parameters and 30 times less distinct VLP images. Without the help of object tags, OSCAR even lags behind our models without VLP.

\textbf{I-Tuning with Dropping.} Since the lower layers of GPT2 have weaker representation ability, we investigate whether we can drop the \textbf{I-Tuning} modules in the first few layers, so that we can further reduce the computational overhead during training and inference. Table~\ref{tab:previous_work_comparsion} shows that it is not necessary to include the \textbf{I-Tuning} models in all layers. Especially, dropping the \textbf{I-Tuning} modules in the first-18 layers can even improve the performance of our model on some evaluation metrics, while the number of trainable parameters is reduced by 50\%, improving the efficiency of our models.

\textbf{Qualitative Evaluation.} Table~\ref{tab:generated_example} presents the image captioning examples of \textbf{I-Tuning}, OSCAR and ClipCap for the first 4 images in the MSCOCO test set. The generated captions of \textbf{I-Tuning} depict the image successfully, which can identify the movement of the people in the image. For example, our model can recognize that the little girl is blowing the candles, while ClipCap and OSCAR cannot.

\subsection{Cross-Attention Visualization}\label{sec:vis}

\begin{figure}
    \centering
    \includegraphics[height=2.2cm]{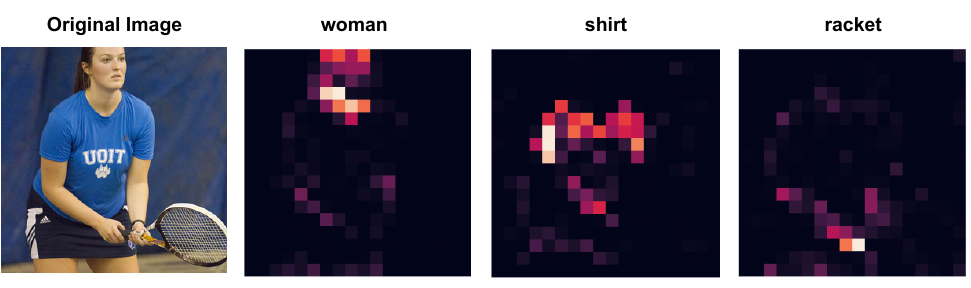}
    \caption{Visualization of the cross-attention maps of text tokens in the caption ``A woman in a blue shirt holding a racket".}
    \label{fig:visualization}
\end{figure}

%In this section, 
We visualize the cross-attention maps of \textbf{I-Tuning} to examine whether it learns the cross-modal information alignment implicitly. We randomly choose an image in the MSCOCO dataset and present the cross-attention heatmaps in the final \textbf{I-Tuning} module of GPT2-Large. Figure~\ref{fig:visualization} shows that our \textbf{I-Tuning} module can correctly attend to the corresponding image regions given different tokens. These examples reveal that our method can learn visual grounding implicitly.

\section{Conclusion}
In this paper, we present a novel lightweight image captioning framework, \textbf{I-Tuning}, which efficiently tunes the frozen PLMs with images. Extensive experiments are conducted to verify the effectiveness of our method. Compared with the baseline systems, our method achieves comparable or even better performance, while our models require up to \textbf{10 times fewer} trainable parameters and \textbf{much fewer} training data.

\section*{Acknowledgment}
This work is partially supported by National Natural Science Foundation of China Young Scientists Fund (No. 62206233), Hong Kong RGC ECS (No. 22200722), and the Key Research and Development Program of Zhejiang Province (No. 2022C01011).

\bibliographystyle{IEEEbib}
\bibliography{strings,refs}

\end{document}